\documentclass{article}
 \usepackage{graphicx}
\usepackage{placeins}
  \usepackage[preprint]{neurips_2019}
\usepackage{subcaption}\def\lone{$\ell_1$-norm~}
\usepackage[utf8]{inputenc}
\usepackage[T1]{fontenc}    
\usepackage{hyperref}      
\usepackage{url}           
\usepackage{booktabs}     
\usepackage{amsfonts}      
\usepackage{nicefrac}       
\usepackage{microtype}      

\renewcommand{\cite}[1]{\citep{#1}}
\usepackage{enumitem}

\title{A Closer Look at Structured Pruning for Neural Network Compression}

\author{
  Elliot J. Crowley*\qquad Jack Turner*\qquad Amos Storkey\qquad Michael O'Boyle\\
  School of Informatics\\
  University of Edinburgh\\
    \texttt{\scriptsize{\{elliot.j.crowley, jack.turner, a.storkey\}@ed.ac.uk}, mob@inf.ed.ac.uk} \\
}

\begin{document}

\maketitle

\begin{abstract}
Structured pruning is a popular method for compressing a neural network: given a large trained network, one alternates between removing channel connections and fine-tuning; reducing the overall width of the network. However, the efficacy of structured pruning has largely evaded scrutiny. In this paper, we examine ResNets and DenseNets obtained through structured pruning-and-tuning and make two interesting observations: (i) {\it reduced networks}---smaller versions of the original network trained from scratch---consistently outperform pruned networks; (ii) if one takes the architecture of a pruned network and then trains it {\it from scratch} it is significantly more competitive. Furthermore, these architectures are easy to approximate: we can prune once and obtain a family of new, scalable network architectures that can simply be trained from scratch. Finally, we compare the inference speed of {\it reduced} and pruned networks on hardware, and show that reduced networks are significantly faster. Code is available at~\url{https://github.com/BayesWatch/pytorch-prunes}.
\end{abstract}

\section{Introduction}

Deep neural networks excel at a multitude of tasks, but are typically cumbersome and difficult to deploy on embedded devices~\cite{sze2017efficient}. This can be rectified by compressing the network; specifically, reducing the number of parameters it uses in order to reduce its dynamic memory footprint. This also reduces the number of operations that have to be performed, making the network faster in theory. A popular means of doing this is through network pruning. Starting with a large trained network, one performs {\it pruning-and-tuning}: alternating between (i) removing connections according to some saliency metric, and (ii) fine-tuning the network.

Pruning methods largely fall into two categories: unstructured (weight pruning) and structured (channel pruning). Although excellent characterisations of the former exist~\cite{frankle2019lottery, gale2019state}, these rely on the future development of specialised hardware capable of leveraging sparse convolutions into real inference speedups. {\bf In this work we focus solely on structured pruning} as this is more applicable to running networks on existing general-purpose hardware.

Now, {\it pruning-and-tuning} is a rather elaborate procedure, can be expensive, and requires one to have---or have to train---the initial network. It would be much simpler to just train a smaller version of this network (e.g.\ one with reduced width or depth) to suit a particular parameter budget. With this in mind, we ask: would these smaller networks outperform their pruned-and-tuned counterparts? If so, why should we perform structured pruning at all?

In this work, we perform \lone channel pruning~\cite{li2016pruning} and Fisher pruning~\cite{theis2018faster,molchanov2017pruning} on modular networks trained on the CIFAR-10~\cite{krizhevsky2009learning} and ImageNet~\cite{russakovsky2015imagenet} datasets. \lone pruning is widely used, and Fisher pruning is a principled, state-of-the-art pruning technique. We make two interesting observations:
\begin{enumerate}

    \item For a given parameter budget, pruned-and-tuned networks are consistently beaten by smaller versions of the original network---or, {\it reduced} networks---trained from scratch. 
    \item  When the architectures obtained through structured pruning are trained {\it from scratch} they surpass their fine-tuned equivalents (and reduced networks in the case of residual nets).
\end{enumerate}

Moreover, we show that the architectures obtained through structured pruning are easy to approximate; we can look at which connections remain after pruning and derive a family of copycat architectures that, when trained from scratch, display similar performance.

First, we outline the pruning procedure, and justify our design choices (Section~\ref{sec:prelim}). Then in Section~\ref{sec:cifar} we prune an exemplar WideResNet and DenseNet on CIFAR-10. We compare these pruned-and-tuned networks to (i) reduced networks, and (ii) pruned networks trained from scratch. In Section~\ref{sec:prunenas} we examine the architectures obtained through pruning and derive a family of copycat networks. We then demonstrate that our pruning observations hold on ImageNet in Section~\ref{sec:additional}.  Finally, in Section~\ref{sec:hardware} we benchmark the inference speed of pruned and reduced networks.

\section{Related work}

Pruning was initially developed as an analogy to finding the minimum description length of neural networks~\cite{hanson1989comparing,lecun1989optimal}. It was later repurposed in several studies~\cite{han2015learning, han2016deep,scardapane2017group, molchanov2017variational} to allow neural networks to be deployed in resource-constrained situations. These methods are all forms of {\it unstructured} pruning. This results in sparse weight matrices, which are difficult to leverage into performance improvements on general purpose hardware~\cite{turner2018characterising}. Even in many specialised cases---such as the popular Tensor Processing Unit~\cite{jouppi2017datacenter}---there exists no in-built support for sparsity. The current state-of-the-art sparse neural network accelerator achieves a peak throughput of 2 tera-ops per second~\cite{parashar2017scnn}, which is two orders of magnitude slower than the Nvidia Volta GPU (125 tera-ops per second). Furthermore, many of these techniques heavily target large fully-connected layers, whose use has subsided in modern network designs~\cite{huang2017densely}. 

More structured approaches have been developed to allow the benefits of pruning to be exposed to general purpose devices~\cite{he2017channel,ye2018rethinking,louizos2017bayesian,liu2017learning}, all of which revolve around removing \textit{entire channels}. By maintaining the density of the weight matrices, these methods are easily leveraged into performance improvements on non-specialised hardware without significant reductions in predictive ability~\cite{molchanov2017pruning,louizos2017bayesian}. However, these pruning methods introduce a hyper-parameter overhead; they often require thresholds, pruning rates, and sensitivity parameters to be hand-tuned. Two automation processes have been explored in the literature: guided search through the space of sensitivity parameters~\cite{yang2018netadapt, he2018amc}, and sensitivity analysis embedded into the saliency estimation~\cite{molchanov2017pruning, theis2018faster}. 

\citet{liu2019rethinking} have also demonstrated the benefits of training pruned models from scratch, as opposed to pruning-and-tuning (although this has been hotly debated in~\citealp{gale2019state} for  {\bf unstructured} pruning scenarios). Our work however, differs in several key aspects:

\begin{enumerate}
    \item We consider scenarios where only inference is resource-constrained, not training. Thus, where~\citet{liu2019rethinking} retrain pruned models with computational budgets, we train to convergence. We believe this is both more representative of the typical pruning workflow and is a fairer comparison.
    \item We compare a higher performance pruning technique~\cite{theis2018faster} and several alternatives to pruning on higher performance models. We also use iterative pruning where~\citet{liu2019rethinking} use one-shot pruning; iterative pruning has been shown to give better performance~\cite{han2015learning}, as well as providing more comparison points.
    \item We additionally examine pruning from an inference time and resource efficiency perspective on platforms that represent typical deployment targets. We show that the width reduction caused by structured pruning (reducing the number of channels per layer) harms the throughput of networks on both CPUs and GPUs and propose that when inference speed is a priority, it is often best to choose reduced-depth models.
\end{enumerate}

\section{Preliminaries}
\label{sec:prelim}

Structured pruning of a neural network broadly corresponds to carrying out the following procedure:
\begin{enumerate}
\item{Train a large neural network from scratch / obtain a pre-trained network off-the-shelf}
\item{Rank candidate channels according to how important they are by some metric $\Delta_{c}$}
\item{Remove the lowest ranked channel(s)}
\item{Fine-tune the network. If the network is the desired size, stop. Otherwise return to Step 2.}
\end{enumerate}

This is most commonly done either sequentially---at Step 3, only remove the single lowest ranked channel---or in a single shot: at Step 3, remove a whole subset of the channels to reduce the network to a specific size.

In this work we focus on sequential pruning because it produces a whole family of networks of different sizes, as opposed to a single network with one-shot pruning, thus providing many more comparison points. It has the added bonus that it would allow for the deployment of pruned networks on multiple devices with different capabilities without knowing exact specifications ahead of time.

\subsection{Pruning modular networks}
\label{sec:pruningmodular}

In this work, we perform structured pruning of residual networks~\cite{he2016deep,zagoruyko2016wide} and DenseNets~\cite{huang2017densely}. These feature modular blocks and skip connections, and are therefore representative of a large number of commonly used modern networks. We avoid older networks such as AlexNet~\cite{krizhevsky2012imagenet} and VGG nets~\cite{simonyan2015deep} as they prove highly redundant~\cite{springenberg2015striving} due to their excessively large fully-connected layers. Here, we describe how such modular networks are pruned.

\begin{figure}[!b]
          \begin{minipage}[b]{0.5\linewidth}
            \centering\includegraphics[width=0.7\textwidth]{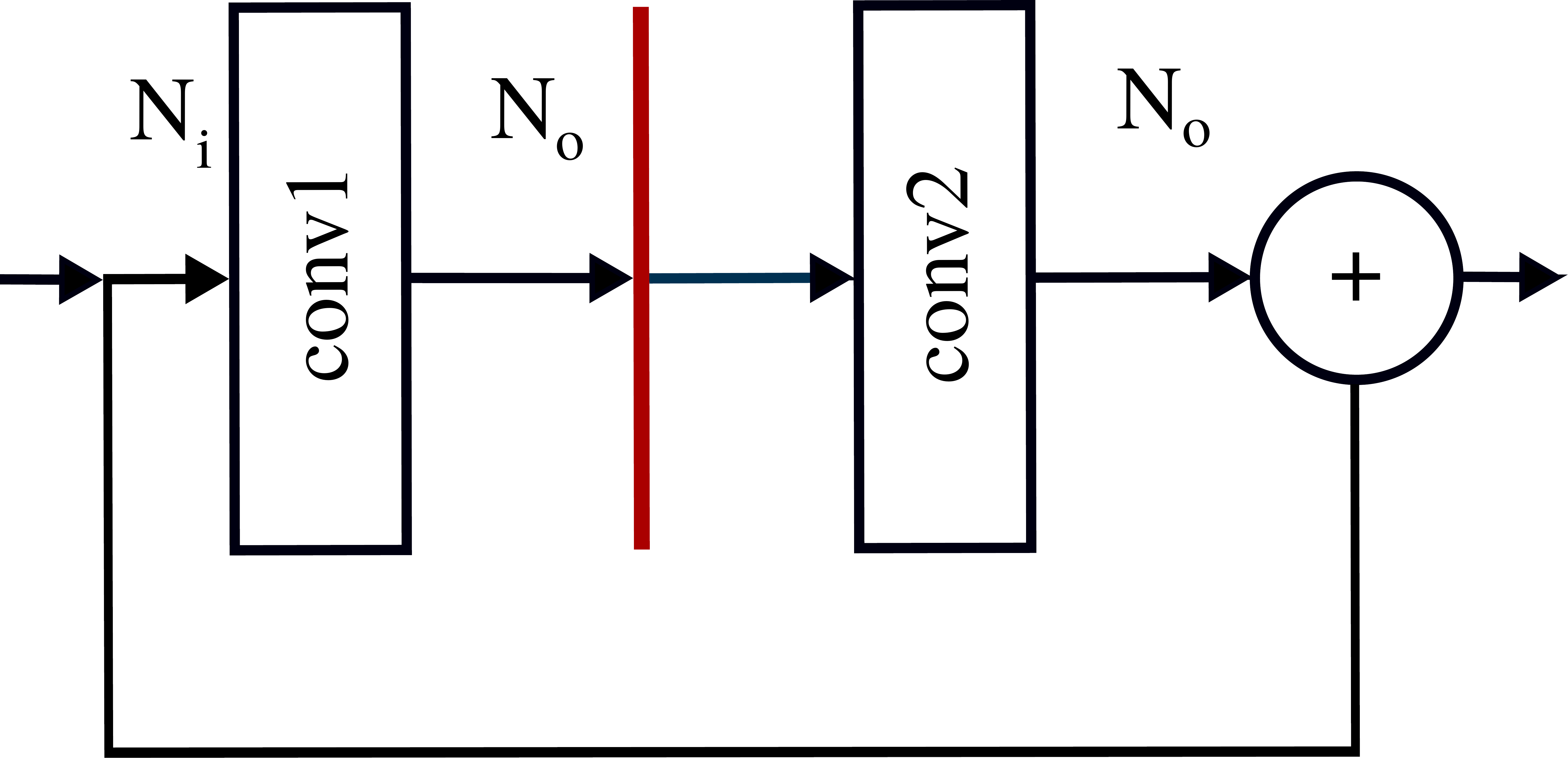}
            \subcaption{}
            \label{blocks:a}
          \end{minipage}
          \begin{minipage}[b]{0.5\linewidth}
            \centering\includegraphics[width=0.7\textwidth]{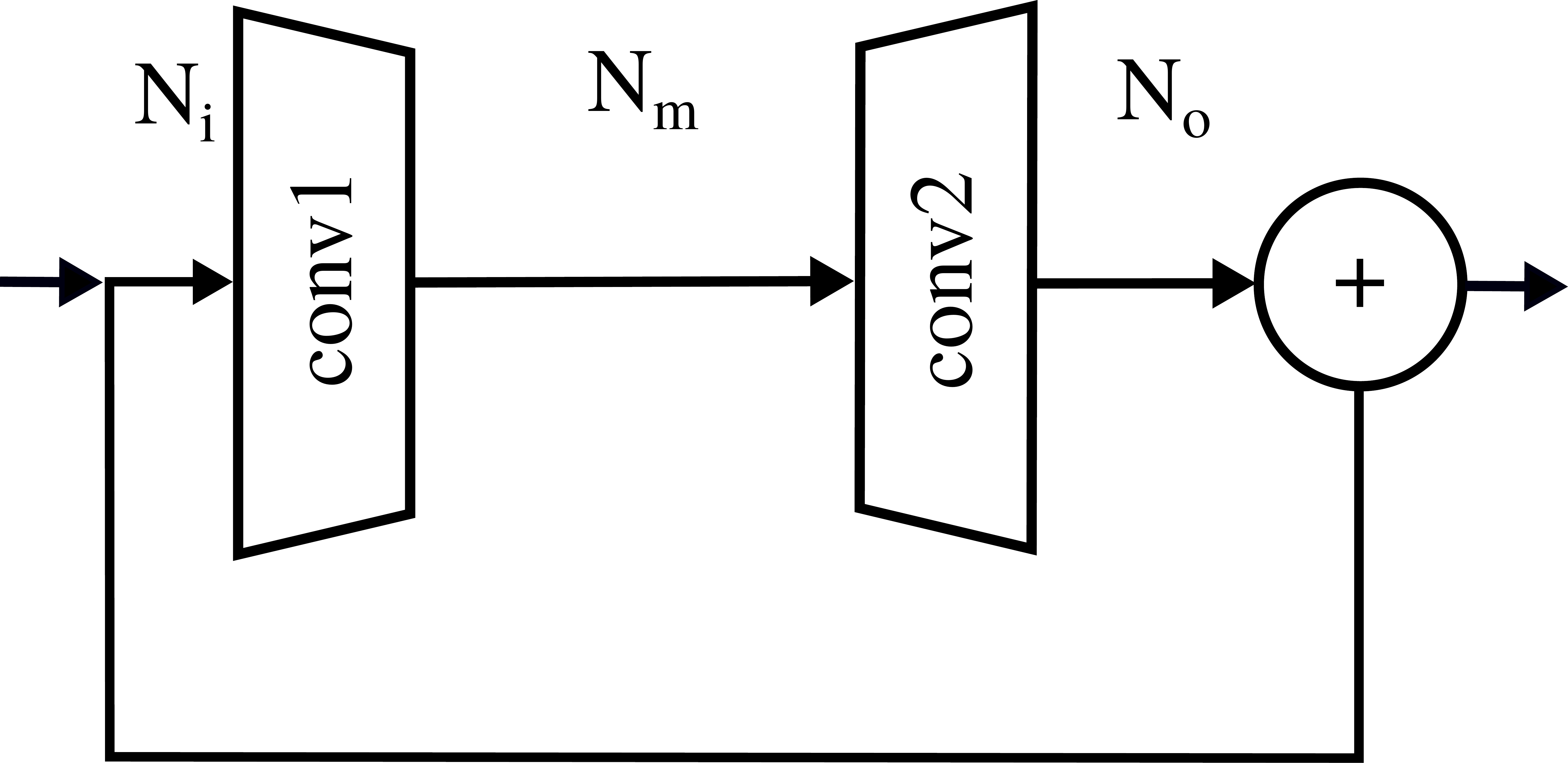}

            \subcaption{}
            \label{blocks:b}
          \end{minipage}
        \caption{({\subref{blocks:a}}) A residual block, consisting of two convolutional layers (the first takes the input from $N_i$ channels to $N_o$ channels, and the second from $N_o$ channels to $N_o$ channels), and a skip connection. In standard residual networks, $N_i \leq N_o$.  Batch-norm + ReLU are omitted for simplicity. The red line is where channels of the activation are pruned. ({\subref{blocks:b}}) A bottlenecked residual block where the intermediate channel dimension is $N_m$ where $N_m <N_o$. Pruning has the effect of transforming ({\subref{blocks:a}}) to ({\subref{blocks:b}}). Within a pruned network the number of intermediate channels remaining $N_m$ will differ from block to block according to which channels are deemed unimportant.}

        \label{fig:pruneblock}
\end{figure}

Consider a standard block in a ResNet (Figure~\ref{blocks:a}) that takes some representation of an image with $N_{i}$ channels and contains two standard convolutional layers. Let us assume that its convolutions use $3 \times 3$ kernels, and the cost of batch-norm~\cite{ioffe2015batch} is negligible. The first convolution changes the number of channels of the representation to $N_{o}$. It consists of $N_{o}$ lots of $N_{i} \times 3 \times 3$ filters and therefore uses $9N_{i}N_{o}$ parameters. The second convolution does not change the number of channels. It consists of $N_{o}$ lots of $N_{o} \times 3 \times 3$ filters and therefore uses $9N_{o}^2$ parameters. We prune this block by removing whole channels of the activation between the two convolutions (see the red vertical line in Figure~\ref{blocks:a}). If we remove $(N_{o}-N_{m})$ of these channels this has the effect of changing the number of intermediate channels to $N_{m}$ (Figure~\ref{blocks:b}). This allows us to throw away filters in the first convolution (which reduces to $N_{m}$ lots of $N_{i} \times 3 \times 3$ filters) and decrease the filter channel size in the second convolution ($N_{o}$ lots of $N_{m} \times 3 \times 3$ filters). This decreases the number of parameters the block uses by a factor of $N_{o}/N_{m}$. Note that when a whole network consisting of $j$ of these blocks is pruned, each block will have a unique $N_{m}$ value: $N_{mj}$.

DenseNets with bottleneck blocks also consist of two convolutional layers, so can be similarly pruned. For a DenseNet with growth rate $k$, the first convolution uses $1 \times 1$ kernels, and reduces the number of input channels from $N_{i}$ to $4k$. The second has $3 \times 3$ kernels, and reduces the number of channels down further to $k$. This output is then concatenated with the input and forms the input to the next block. We can again zero out channels of the intermediate activation to compress both convolutions.

\subsection{\lone based pruning}

We utilise the method of~\citet{li2016pruning}. They show that removing channel activations whose corresponding filter's weights have the smallest absolute sum is superior to random or largest-sum pruning: Consider a weight tensor consisting of $N_{o}$ lots of $N_{i}\times K \times K$ filters. Each filter is applied to some $N_{i}$ channel input activation, producing one of $N_{o}$ output activation channels. The importance metric $\Delta_{c}$ for each of these output channels is just the sum of each of these filters. We can then rank every channel in the model by $\Delta_{c}$, and prune the ones with the smallest $t$ norms, where $t$ is a temperature parameter that controls the number of channels pruned at each pruning stage. We use $t=1$ and prune iteratively, as~\citet{li2016pruning} reported that this results in the smallest increase in error. From here onwards we refer to this method as \lone pruning.

\subsection{Fisher pruning}

In Fisher pruning~\citep{theis2018faster,molchanov2017pruning}, channels are ranked according to the estimated change in loss that would occur on their removal (i.e.\ this quantity is used as $\Delta_{c}$). Consider a single channel of an activation in a network due to some input minibatch of $N$ examples. Let us denote this as $C$: it is an $N \times W \times H$ tensor where $W$ and $H$ are the channel's spatial width and height. Let us refer to the entry corresponding to example $n$ in the mini-batch at location $(i,j)$ as $C_{nij}$. If the network has a loss function $\mathcal{L}$, then we can back-propagate to get the gradient of the loss with respect to this activation channel $\frac{\partial \mathcal{L}}{\partial {C}}$. Let us denote this gradient as $g$ and index it as $g_{nij}$. $\Delta_{c}$ can then be approximated as: 
\vspace{-2mm}
\begin{equation}
 \Delta_{c} =\frac{1}{2N} \sum_{n}^{N}\left(- \sum_{i}^{W}\sum_{j}^{H}A_{nij} g_{nij}\right)^2, \label{eqn:fisher}
\end{equation}

The derivation for this can be found in \citet{theis2018faster}. Note that their $\Delta_{c}$ is slightly different from that in~\citet{molchanov2017pruning}, although we found there to be little difference in practice.
\section{CIFAR-10 experiments}
\label{sec:cifar}

We perform a set of extensive experiments on an exemplar WideResNet and DenseNet, representatives of two popular network types, to ascertain whether the structured pruning procedure is worthwhile over some simpler process. Specifically, in Section~\ref{sec:prunevscratch} we compare pruning-and-tuning these networks to the simple alternative of training a reduced version of the original network from scratch (e.g.\ one with lower depth). We refer to the networks obtained this way as {\it reduced} networks. In Section~\ref{sec:prunevscratch} we compare pruned-and-tuned networks to those with the {\it same architecture} but trained from scratch. We evaluate the performance of our networks by their classification error rate on the test set of CIFAR-10~\citep{krizhevsky2009learning}. Implementation details are given at the end of the section.

\subsection{Pruned networks vs. reduced networks}
\label{sec:prunevscratch}

\begin{figure}[!p]
          \begin{minipage}[]{\linewidth}
            \centering\includegraphics[width=0.98\textwidth]{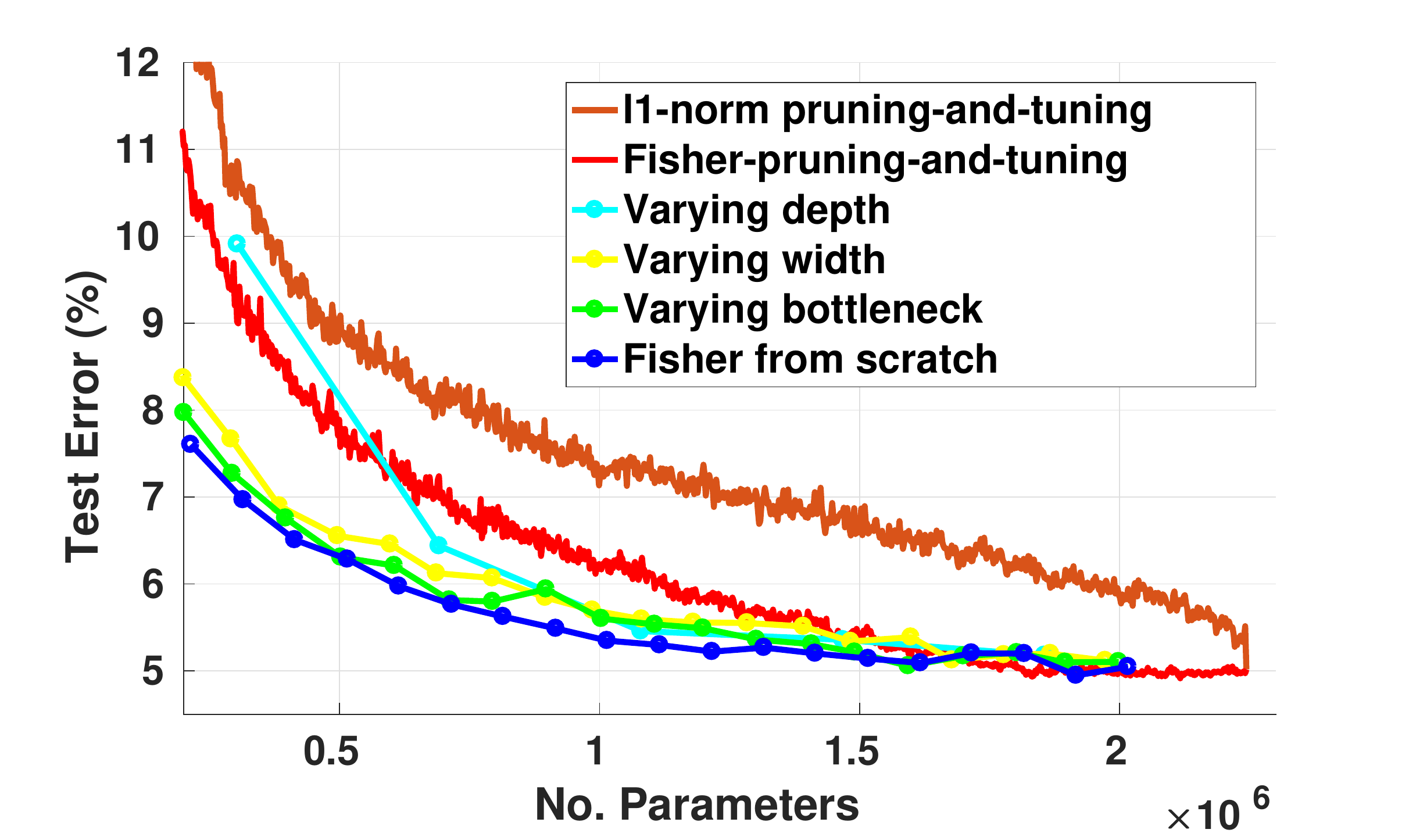}
            \subcaption{}
            \label{traj:a}
            \end{minipage}
          \begin{minipage}[]{\linewidth}
            \centering\includegraphics[width=0.98\textwidth]{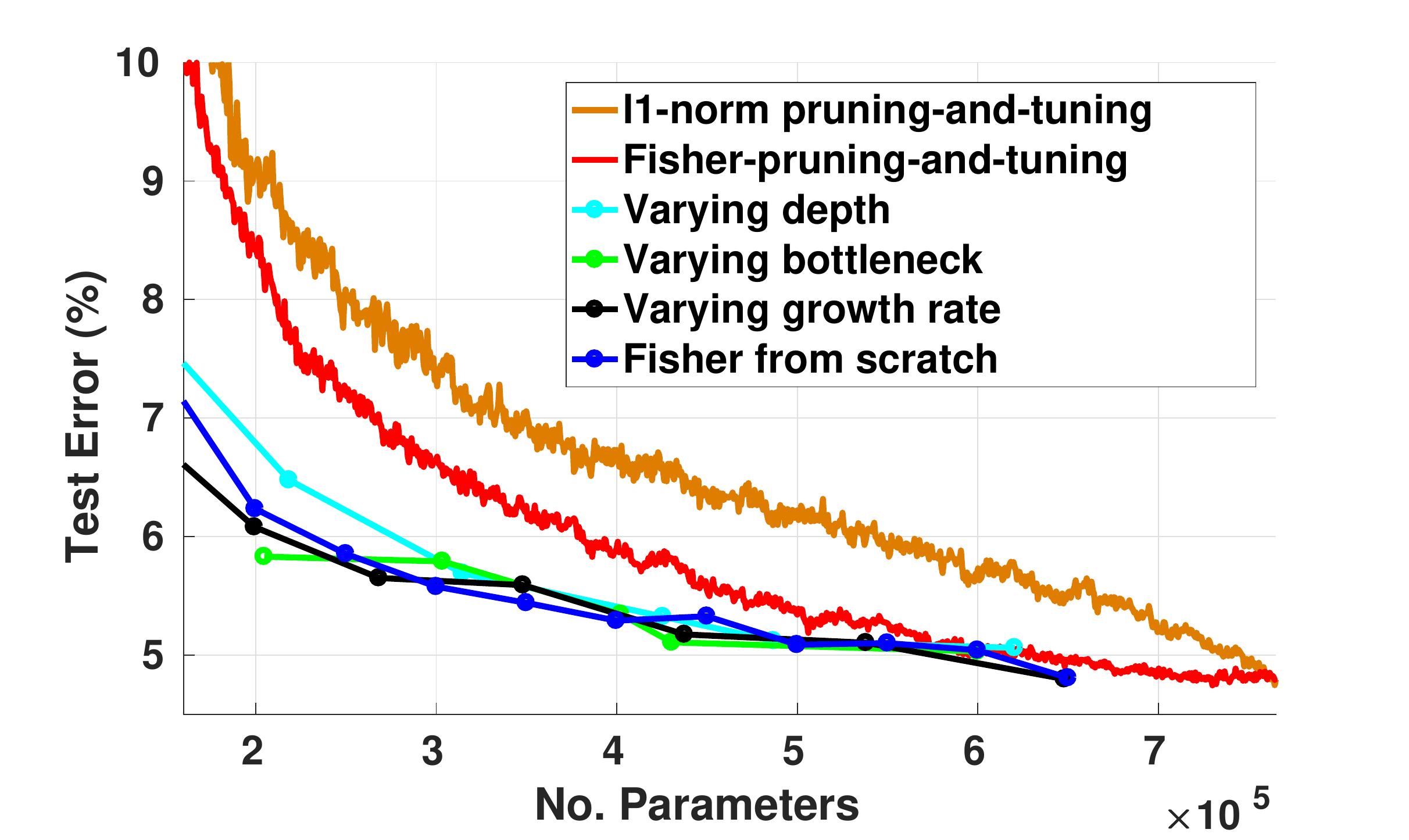}
            \subcaption{}
            \label{traj:b}
            \end{minipage}

        \caption{Test Error~vs.~No.~Parameters for ({\subref{traj:a}}) WRN-40-2 and ({\subref{traj:b}}) DenseNet-BC-100 ($k$=12) variants on CIFAR-10. The orange curve corresponds to the original network undergoing \lone pruning and fine-tuning, and the red curve is Fisher pruning-and-tuning. Notice that Fisher pruning is a better pruning technique in both cases. We compare these pruned-and-tuned networks to reduced versions of the original networks trained from scratch. In ({\subref{traj:a}}) the WideResNet is reduced by either varying its depth (cyan curve), width (yellow), or the bottlenecks in its blocks (green). In ({\subref{traj:b}}) the DenseNet is reduced by varying its  depth (cyan), growth rate (black), or bottlenecks (green). {\bf These reduced networks largely outperform pruned-and-tuned networks across most parameter budgets}. If we take Fisher-pruned architectures and train them from scratch we get the blue curve; for WideResNets, networks along this curve tend to surpass all others. Each circle corresponds to a trained network with test error averaged over 3 runs. Pruning curves displayed are averaged across 5 independent pruning runs for 5 different trained models. The architectures used to train the models on the blue curve were taken from a randomly selected pruning run.}
\label{fig:prunecurves}
\end{figure}

\subsubsection{WideResNet}
For our exemplar WideResNet~\citep{zagoruyko2016wide} we use WRN-40-2, i.e.\ a WideResNet with 40 layers and a channel width multiplier of 2. It has 2.2M parameters in total, the bulk of which lie in 18 residual blocks, each containing two convolutional layers (Figure~\ref{blocks:a}). The network is first trained from scratch, and is then separately \lone pruned and Fisher-pruned, as described in Section~\ref{sec:prelim}. In each case, before a channel is removed, the test error and parameter count of the network is recorded. The resulting trajectory of Test Error versus Number of Parameters is represented by the orange curve for \lone pruning in Figure~\ref{traj:a}, and the red curve for Fisher pruning. Notice that Fisher pruning greatly outperforms \lone pruning; moreover, \lone pruning has an immediate harmful impact on the network's error.

We compare this process to training {\it reduced} WideResNets from scratch. We train WRN-40-$w$ networks---the yellow curve in Figure~\ref{traj:a}---varying non-integer $w$ (and therefore network width) to produce networks that have between 200K and 2M parameters inclusive, in 100K intervals. We also train WideResNets with reduced depth (the cyan curve); specifically WRN-$d$-2 for $d$ between 10 and 34, in intervals of 6. Finally, we train WRN-40-2 networks where we introduce uniform bottlenecks by applying a variable scalar multiplier $z < 1$ to the middle channel dimension (the red line in Figure~\ref{blocks:a}) in each block. In detail, if we index each block by $j$, then its middle channel dimension changes from its original number, $N_{oj}$ to $zN_{oj}$. This gives the green curve in Figure~\ref{traj:a}. Note that in Figure 2, the circles along each curve correspond to the actual trained networks.

We can see that these reduced networks trained from scratch outperform the pruned-and-tuned networks for most parameter budgets; an exception being when the reduced network is too shallow~\cite{ba2014do}. Of these non-exhaustive families of reduced networks, those with bottlenecks appear to be the most performant. Interestingly, this is the technique that closest resembles pruning. The performance gap is more pronounced the smaller the networks get: above 1.7 million parameters, the Fisher pruned-and-tuned networks are roughly on par with their reduced counterparts, although below this the gap grows significantly. This can be seen in Table~\ref{table:budgets} where we compare the networks across three evenly-spaced parameter budgets (500K, 1M, and 1.5M parameters) which roughly correspond to a 75, 50, and 25\% compression rate of the original network. The number of multiply-accumulate operations (MACs) each network uses is given for completeness.

\subsubsection{DenseNet}

Our exemplar DenseNet (DenseNet-BC-100)~\cite{huang2017densely} uses bottleneck blocks, has a depth of 100, a growth rate of 12, and a transition rate of 0.5. It is already very efficient at $\sim$0.8M parameters. \lone and Fisher pruning-and-tuning this network gives the orange and red curves respectively in Figure~\ref{traj:b}. Notice that again, Fisher pruning substantially outperforms \lone pruning. To produce reduced networks, we vary (i) the growth rate, (ii) depth, and (iii) bottlenecks of the original network. These are given by the black, cyan, and green curves in Figure~\ref{traj:b}, in which circles correspond to the actual trained networks. Once again, the reduced networks outperform the pruned-and-tuned networks for most parameter budgets.

\subsection{Pruned networks trained from scratch}
\label{sec:prunefromscratch}

\begin{table*}[t!]
    \caption{Test errors (\%) on CIFAR-10 for WRN-40-2 variants at three approximate parameter budgets (500K, 1M, 1.5M). In some cases it was not possible to get a network with the exact number of parameters (e.g. there is no WRN-$d$-2 with integer $d$ that has 500K parameters) so actual parameter totals are given (in millions) for clarity. The number of multiple-accumulates (MACs) each network uses is also given (in millions). Note that Fisher pruning outperforms \lone pruning, and as the parameter budget gets lower, the gap between networks trained from scratch and pruned-and-tuned networks markedly increases. Test errors for models trained from scratch are the average across 3 training runs. Errors for pruned-and-tuned models correspond to a randomly selected run.}

        \vskip 0.15in
    \centering
    \scalebox{0.80}{\begin{tabular}{@{}ll|rrr|rrr|rrr@{}}
              \multicolumn{2}{c}{}&\multicolumn{3}{c}{500K Budget} &\multicolumn{3}{c}{1M Budget}&\multicolumn{3}{c}{1.5M Budget} \\
        \toprule
        Method & Scratch?&Params & MACs &Error& Params & MACs  &Error& Params & MACs  &Error\\ 
        \hline
        \lone pruning&$\times$ &0.51 &83 & 9.14&1.02&168&7.39&1.52&254&6.37\\
         Fisher pruning&$\times$ & 0.52& 88& 7.41&1.02&173&6.49&1.52&247&5.49\\
         Varying Depth&\checkmark& 0.69& 101& 6.44 &1.08&158& 5.46&1.47&215&5.36\\
         Varying Width &\checkmark   & 0.50 & 74 & 6.55&0.98&144&5.70&1.48&218&5.34\\
         Varying Bottleneck &\checkmark       &0.50&74 &6.31 & 1.00 & 146&5.60&1.49&217&5.21\\
         Fisher Scratch &\checkmark                        & 0.52& 88& 6.28&1.02&173&5.35&1.52&247&5.14\\

        \bottomrule
    \end{tabular}}

    \label{table:budgets}

\end{table*}

In Section~\ref{sec:prunevscratch}, pruned-and-tuned WideResNets and DenseNets were consistently beaten by reduced networks trained from scratch. How are these pruned-and-tuned networks losing to such a simple alternative? This could be due to the regularisation introduced by long periods of training~\citep{hoffer2017train} or the importance of the early phases of learning~\cite{jastrzkebski2019relation,achille2019critical}, necessitating that the network should be trained from scratch rather than fine-tuned~\cite{he2018rethinking}. If this is the case then there may yet be a use for the {\it architectures} found through pruning.

To ascertain this, we take WideResNet architectures obtained through a randomly selected Fisher pruning run. We train these for a range of parameter totals (roughly 200K to 2M inclusive in 100K intervals), reinitialise the weights at random and train anew. This gives the blue curve in Figure~\ref{traj:a}: notice that these networks are significantly better than those obtained through Fisher pruning-and-tuning (the red curve). Furthermore, these networks largely outperform the {\it reduced} WideResNets as well; this can be observed for specific parameter budgets in Table~\ref{table:budgets}. These results support the case for pruning being a form of architecture search~\cite{frankle2019lottery,liu2019rethinking}.

When we train Fisher-pruned DenseNet architectures in a similar manner we obtain the blue curve in Figure~\ref{traj:b}. Although these similarly outperform their pruned-and-tuned counterparts (red curve) they do {\bf not} outperform the reduced networks; instead, they are roughly on par with them. We hypothesise that this is due to the robust nature of DenseNets: as they encourage feature reuse, the saliency of a particular connection is reduced, such that its removal is of less consequence.

\paragraph{Implementation details} 

Networks are trained from scratch using SGD with momentum to minimise the cross-entropy loss. Images are augmented using horizontal flips and random crops. We use the learning schedules and hyperparameters from~\citet{zagoruyko2016wide} for WideResNets, and~\citet{huang2017densely} for DenseNets. Networks are fine-tuned using the lowest learning rate reached during training. The connection with the lowest $\Delta_{c}$ (see Section~\ref{sec:prelim}) is removed every 100 steps of fine-tuning. We experimented with several different values for the FLOP penalty hyperparameter for Fisher pruning (see~\citealp{theis2018faster}) but found this made little difference. We therefore set it to 0.

\section{Deriving architectures from pruning}
\label{sec:prunenas}

In Section~\ref{sec:prunefromscratch} we observed that Fisher-pruned WideResNets trained {\it from scratch} not only outperformed their pruned-and-tuned counterparts but also {\it reduced} for a range of parameters. This indicates that these pruned networks have useful architectures. We examine the channel profile of a Fisher-pruned WideResNet and derive a new family of scalable reduced networks.

\begin{figure}[b!]
          \begin{minipage}[b]{0.5\linewidth}
            \centering\includegraphics[width=.8\textwidth]{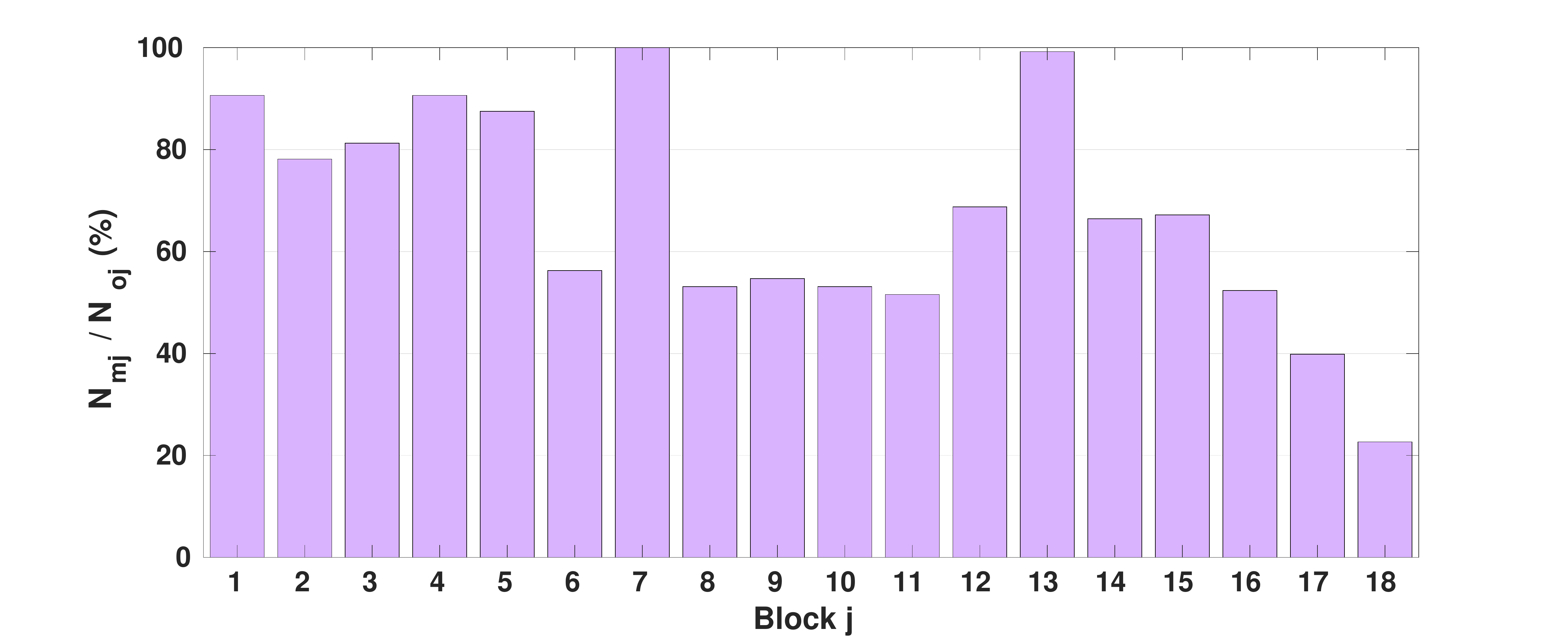}
            \subcaption{}
            \label{profile:a}
          \end{minipage}
          \begin{minipage}[b]{0.5\linewidth}
            \centering\includegraphics[width=.8\textwidth]{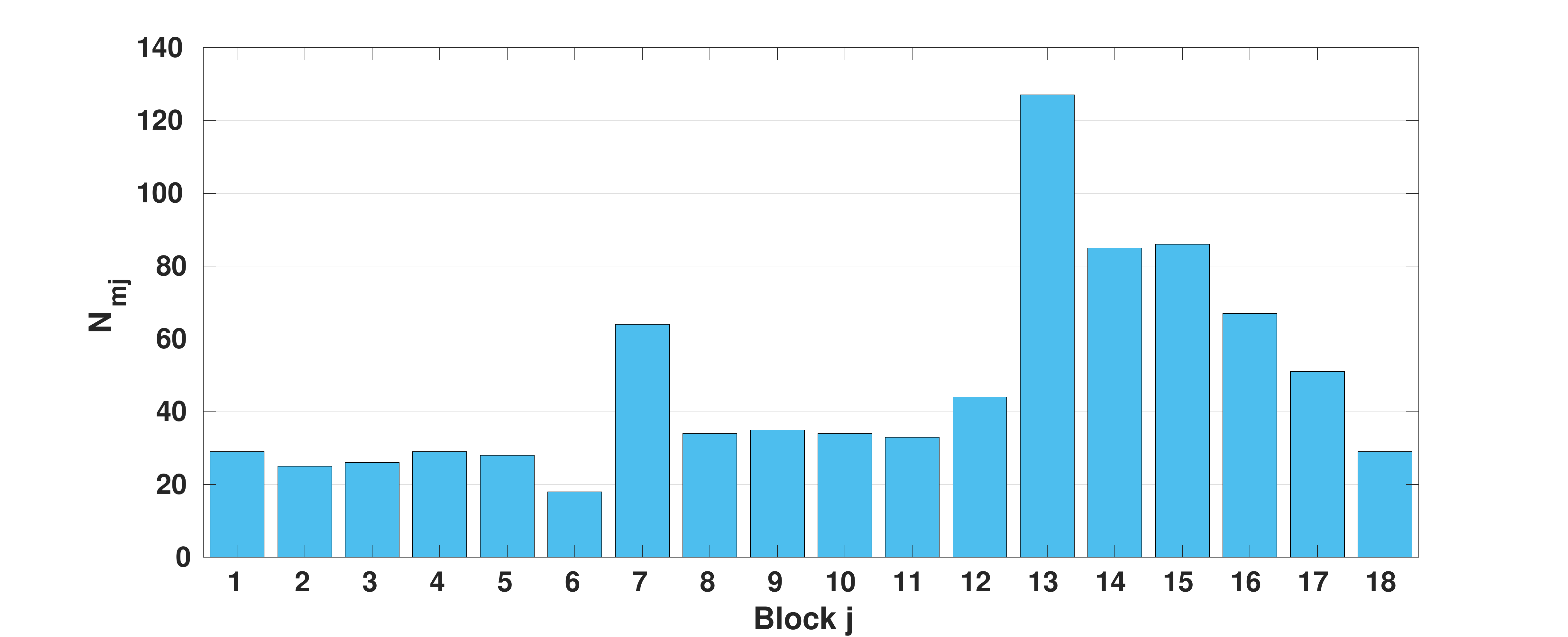}

            \subcaption{}
            \label{profile:b}
          \end{minipage}
          \begin{minipage}[b]{\linewidth}
        \centering\includegraphics[width=.7\textwidth]{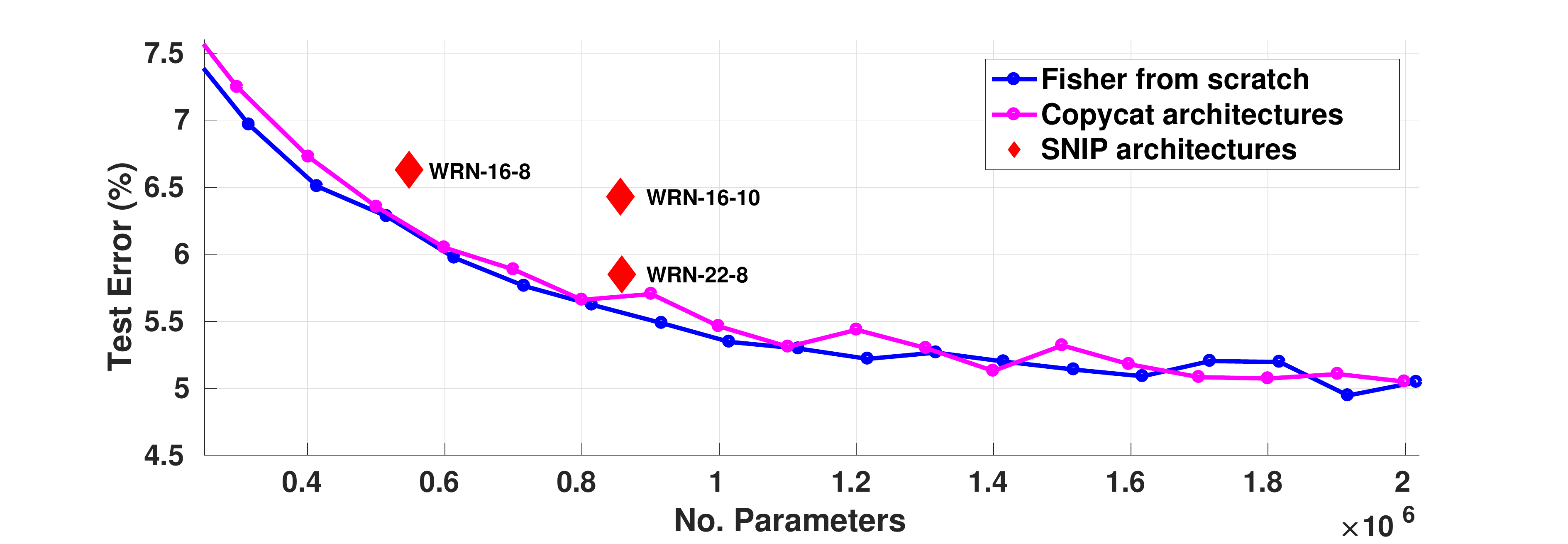}
            \subcaption{}
            \label{fig:rescopy}
                      \end{minipage}

\caption{{({\subref{profile:a}}) shows the percentage of intermediate channels remaining in each residual block $N_{m}/N_{o}$ of a Fisher-pruned WRN-40-2 after 500
channels have been pruned. The actual number $N_{m}$ is shown in ({\subref{profile:b}}). We linearly scale this profile to produce copycat networks. These are compared to the networks they are mimicking in ({\subref{fig:rescopy}}) (where pink is copycat, and blue is Fisher-pruned from scratch). Errors are the average of three runs. Notice that these copycat networks closely emulate the performance of the Fisher-pruned networks. WideResNets compressed using SNIP~\cite{lee2019snip} are given for comparison (the red diamonds); the copycat networks outperform these networks.
}}
        \label{fig:prune}
\end{figure}

Recall that pruning has the effect of reducing the number of intermediate channels in each block $j$---which we denote as $N_{oj}$---to a unique value $N_{mj}$. For blocks with important connections, the ratio of channels remaining after pruning---$N_{mj}/N_{oj}$---will be high as few channels will have been pruned, whereas for blocks with expendable connections this ratio will be low. Figure~\ref{profile:a} gives this ratio for each block in our WideResNet after pruning-and-tuning away 500 connections. The actual number of intermediate channels $N_{mj}$ is given in Figure~\ref{profile:b},

Notice that channels in later layers tend to be pruned; they are more expendable than those in earlier layers. It is intriguing that channels are rarely pruned in blocks {7 and 13}; these blocks both contain a strided convolution that reduces the spatial resolution of the image representation. It is imperative that information capacity is retained for this process. These observations are consistent with those made in previous works~\citep{veit2016residual,huang2016deep,huang2018condensenet,jastrzkebski2018residual}.

To assess the merits of this particular structure, we train {\it copycat Fisher architectures}, and summarise them as a scalable family of reduced networks designed to mimic those obtained through Fisher pruning. Specifically, we train a range of architectures from scratch where we set the intermediate channel dimension in each block $j$ to $\alpha N_{mj}$ for varying $\alpha$; i.e.\ we are scaling the profile in Figure~\ref{profile:b}.

The copycat networks are represented by the pink curve in Figure~\ref{fig:rescopy}, which we compare directly to the networks they are trying to emulate (those Fisher-pruned from scratch which are represented by the blue curve). For most parameter budgets these networks perform similarly to those found through Fisher pruning; this means we can prune a network once to find a powerful, scalable reduced network architecture. Furthermore, the resulting networks are competitive; they are compared to the WideResNets produced by the SNIP pruning method~\cite{lee2019snip}, represented by the red diamonds in Figure~\ref{fig:rescopy}. The copycat networks outperform these for a comparable number of parameters. Can we use these copycat networks found on CIFAR-10 as a plugin replacement for another problem? We briefly examine this by training networks on the CINIC-10~\cite{darlow2018cinic} dataset: an amalgamation of CIFAR and downsampled ImageNet. A standard reduced network, WRN-16-2 (700K parameters) achieves 16.12\% mean validation error over three runs. A 700K copycat network achieves 15.62\% mean error, indicating that these networks are versatile.

\section{ImageNet experiments}
\label{sec:additional}

In this section, we verify that our structured pruning observations for CIFAR-10 (Section~\ref{sec:cifar}) hold for the more challenging ImageNet dataset~\cite{russakovsky2015imagenet}.

\begin{figure*}[!hb]
        \centering
        \includegraphics[width=.9\textwidth]{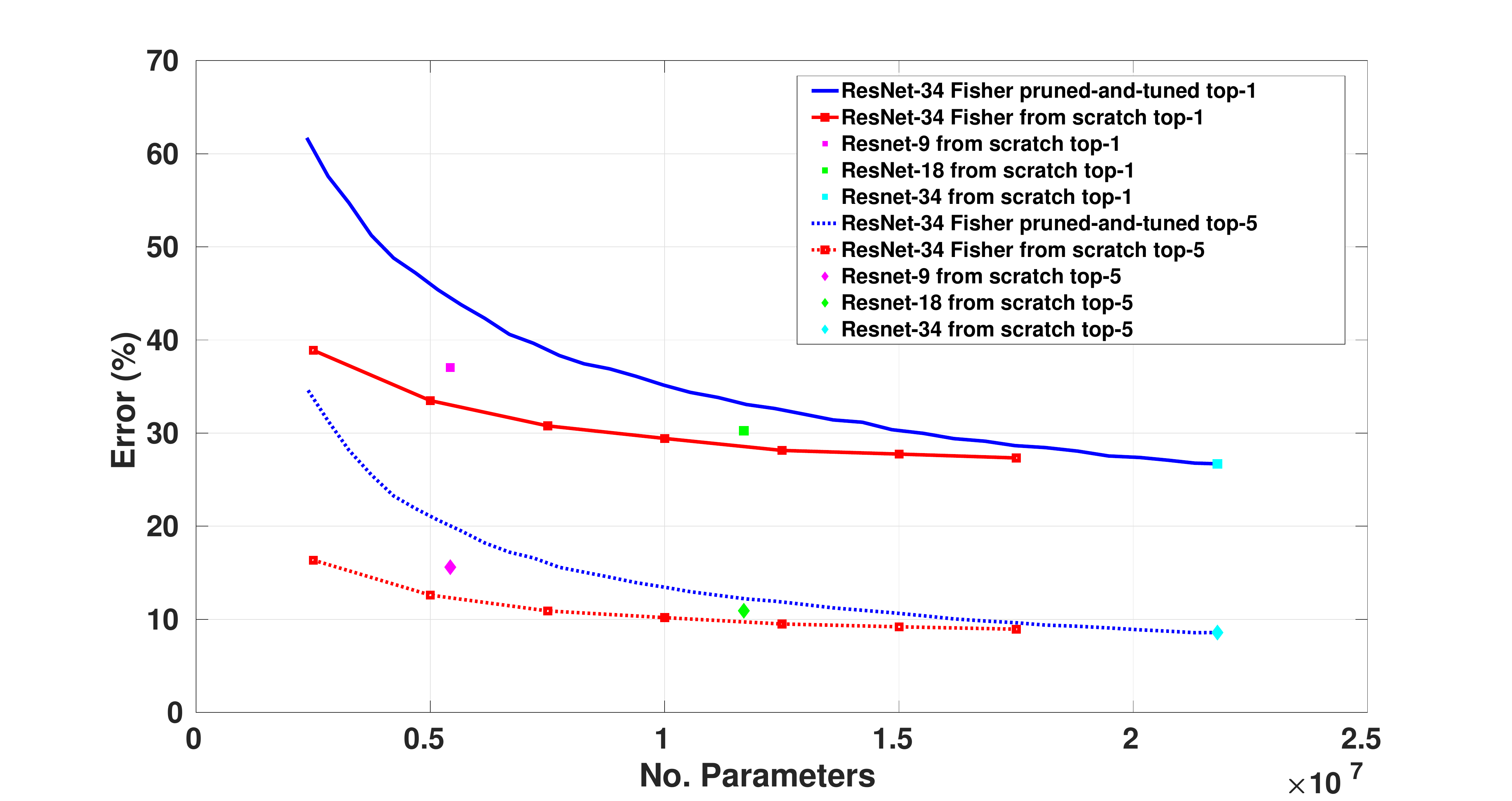}
        \caption{Validation Error~vs.~No.~Parameters for networks on ImageNet. Our initial network---a ResNet-34---has its Top-1 and Top-5 errors denoted by the cyan square and diamond respectively. When we perform Fisher pruning and fine-tuning we obtain the blue curves. Notice that reduced networks---ResNet-9 (pink square/diamond) and ResNet-18 (green square/diamond)---trained from scratch outperform these pruned-and-tuned networks. However, when the pruned architectures are trained from scratch we obtain the red curves which outperform the reduced networks.}
\label{fig:imagenet}

\end{figure*}

We take a trained ResNet-34 and perform Fisher pruning, giving the blue curves in Figure~\ref{fig:imagenet}. The full blue curve corresponds to Top-1 error on the validation set of ImageNet, and the dashed blue curve corresponds to Top-5 error. For our reduced networks we simply train ResNets with lower depth from scratch. The pink square and diamond on Figure~\ref{fig:imagenet} correspond to the top-1 and top-5 error respectively for a ResNet-9. The green square/diamond are the top-1/top-5 error rate for ResNet-18. Notice that for comparable parameter budgets, these reduced networks outperform pruned-and-tuned ResNet-34. We then take networks obtained through Fisher pruning---specifically, those between 2.5 million and 17.5 million parameters inclusive in intervals of 2.5 million---and train them from scratch. This gives us the red curves in Figure~\ref{fig:imagenet} (the circles correspond to the actual trained networks). Notice that the networks on the red curve outperform their blue counterparts, {\it and} the reduced networks (the squares and diamonds). These results are consistent with our previous observations.

\paragraph{Implementation details} Networks are trained from scratch with SGD + momentum for 90 epochs using a minibatch size of 256 split across 4 GPUs with standard crop/flip data augmentation. The initial learning rate was set to 0.1 and was reduced by a factor of 10 every 30 epochs. Momentum was set to 0.9, and weight decay to 0.0001. For Fisher pruning, a connection was removed every 100 steps. The network was fine-tuned with a learning rate of 0.0001. Pre-trained ResNet-34/18 models were obtained from~\url{https://github.com/pytorch/vision}.

\section{Inference acceleration experiments: pruned vs. reduced models}

\label{sec:hardware}

\begin{table*}[b!]
    \caption{A comparison of pruned and reduced models from the hardware performance perspective. Speed is inference time measured in seconds for a single image, MACs/ps represents throughput. We take two reduced models, ResNet-9 and ResNet-18, and compare them to ResNet-34s that have been Fisher-pruned to (a) a similar parameter count, and (b) a similar error level. ResNet-A is chosen to have a similar number of parameters to ResNet-18, while ResNet-B has a similar error. For ResNet-9, ResNet-C has a similar number of parameters and ResNet-D has a similar error. These results indicate that for a given error, {\bf reduced models have faster inference times than pruned models.}}

    \centering
    \scalebox{0.80}{ \begin{tabular}{@{}l|rrr|rr|rr@{}}
        \multicolumn{4}{c}{}&\multicolumn{2}{c}{Core i7 (CPU)} &\multicolumn{2}{c}{1080Ti (GPU)} \\
        \toprule
        Network                    & Params & MACs    & Error   &   Speed  & MACs/ps  & Speed & MACs/ps \\ 
        \hline         ResNet-18     & 11.6M  & 1.81G   & 30.24           &   \textbf{0.060s}    &  3.03      & \textbf{0.002s} & 101.3    \\
         ResNet-34-A   & 12.5M  & 2.42G   & \textbf{28.14}  &   0.085s             &  2.83      & 0.004s & 72.0    \\
         ResNet-34-B   & 7.5M   & 1.78G   & 30.77           &   0.066s             &  2.71      & 0.003s & 49.6   \\
         \hline
         ResNet-9      & 5.4M  & 0.89G   & 37.04             &   \textbf{0.035s}  &  2.52      & \textbf{0.001s} & 79.0   \\
         ResNet-34-C   & 4.9M   & 1.22G   & \textbf{33.49}  &   0.054s           &  2.28      & 0.003s & 35.4     \\
         ResNet-34-D   & 2.5M   & 0.66G   & 38.88           &   0.042s           &  1.16      & 0.003s & 18.0     \\
        \bottomrule
    \end{tabular}}

    \label{table:speed-test}

\end{table*}

We now turn our attention to the effects of structured pruning on \textit{hardware} performance and compare pruned networks to our reduced choices. The most common justification for pruning neural networks is that smaller networks tend to have faster inference times and require less energy to use. There are several papers that report on inference acceleration via pruning networks~\citep{han2016deep,molchanov2017pruning,louizos2017bayesian}; however, pruning is not often compared to simpler approaches. From a resource efficiency perspective, pruning presents a challenge because of the reduced width of each layer. The number of operations performed per parameter is reduced, thereby reducing the opportunities for~\textit{data reuse} in the network. Because of this, it may be preferable to choose models that maintain the width of the original network, reducing the~\textit{depth} instead.
To investigate, we benchmark several ResNet-34 reductions from Section~\ref{sec:additional} using both strategies on a CPU (Intel Core i7) and a GPU (Nvidia 1080Ti). These devices are representative of typical deployment platforms (the GPU for large scale inference in datacentres, the CPU as an example of desktop compute). 

In Table~\ref{table:speed-test}, we explore reductions of ResNet-34 when performing a single inference on ImageNet images. ResNet-18 is compared to two Fisher-pruned ResNet-34s trained from scratch. The first, ResNet-34-A, is chosen to have a similar number of parameters to ResNet-18, and although it has a lower error rate it has a significantly slower inference speed (roughly 33\% slower). ResNet-34-B is chosen to have a similar error rate to ResNet-18, and despite having fewer parameters, it also has a slower inference speed. Note that the ratio of parameters to MACs is extremely poor in ResNet-34-B; this is exemplary of the negative effect that pruning can have on resource efficiency since this pruned model suffers from both high error and slow inference time. We also compare a ResNet-9 to ResNet-34-C (Fisher-pruned to a similar no. of parameters) and ResNet-34-D (Fisher-pruned to a similar error). The same trends appear; the ResNet-34 with the same number of parameters has a lower error but significantly slower inference time, while the ResNet-34 with the same error has fewer parameters but still slower inference. This advantage is highlighted by the MACs/ps performance of the reduced models being higher than the pruned models. We therefore conclude that structured pruning harms the~\textit{throughput} of large neural networks on both CPUs and GPUs. This means that when inference time is a priority, it is best to choose models with reduced depth and train them from scratch.

\section{Conclusion}

We have shown that large pruned-and-tuned networks are consistently beaten by reduced networks for the case of structured pruning. The families of reduced networks used in this work are in no way exhaustive; incorporating grouped convolutions~\cite{ioannou2017deep,huang2018condensenet}, low-rank substitutions~\cite{gray2019separable}, or dilated kernels~\cite{dilated2016} could increase this performance gap even further. Furthermore, we have shown that the network structures produced by structured pruning are not ideally suited to providing fast inferences. However, we have only considered image classification scenarios with large amounts of training data. In transfer learning scenarios moving to a low-data regime, training from scratch may not be a possibility and pruning could prove preferable. Additionally, our experiments with copycat architectures support the notion that pruning is a form of architecture search~\cite{frankle2019lottery}; this top-down form of search benefits from arguably being more intuitive than bottom-up alternatives~\cite{liu2019darts,zoph2018learning}.

\paragraph{Acknowledgements.}
This work was supported in part by the EPSRC Centre for Doctoral Training in Pervasive Parallelism, and funding from the European Union's Horizon 2020
research and innovation programme under grant agreement No. 732204 (Bonseyes).
This work is supported by the Swiss State Secretariat for Education, Research
and Innovation (SERI) under contract number 16.0159. The opinions expressed and
arguments employed herein do not necessarily reflect the official views of
these funding bodies. The authors are grateful to Luke Darlow, David Sterratt, Stanis{\l}aw Jastrz{\k{e}}bski, and Joseph Mellor for their helpful suggestions.

\FloatBarrier
\bibliography{mybib}

\begin{thebibliography}{47}
\providecommand{\natexlab}[1]{#1}
\providecommand{\url}[1]{\texttt{#1}}
\expandafter\ifx\csname urlstyle\endcsname\relax
  \providecommand{\doi}[1]{doi: #1}\else
  \providecommand{\doi}{doi: \begingroup \urlstyle{rm}\Url}\fi

\bibitem[Achille et~al.(2019)Achille, Rovere, and Soatto]{achille2019critical}
Achille, A., Rovere, M., and Soatto, S.
\newblock Critical learning periods in deep neural networks.
\newblock In \emph{International Conference on Learning Representations}, 2019.

\bibitem[Ba \& Caruana(2014)Ba and Caruana]{ba2014do}
Ba, L.~J. and Caruana, R.
\newblock Do deep nets really need to be deep?
\newblock In \emph{Advances in Neural Information Processing Systems}, 2014.

\bibitem[Darlow et~al.(2018)Darlow, Crowley, Antoniou, and
  Storkey]{darlow2018cinic}
Darlow, L.~N., Crowley, E.~J., Antoniou, A., and Storkey, A.
\newblock {CINIC}-10 is not {ImageNet} or {CIFAR}-10.
\newblock \emph{arXiv preprint arXiv:1810.03505}, 2018.

\bibitem[{Frankle} \& {Carbin}(2019){Frankle} and {Carbin}]{frankle2019lottery}
{Frankle}, J. and {Carbin}, M.
\newblock The lottery ticket hypothesis: Finding small, trainable neural
  networks.
\newblock In \emph{International Conference on Learning Representations}, 2019.

\bibitem[Gale et~al.(2019)Gale, Elsen, and Hooker]{gale2019state}
Gale, T., Elsen, E., and Hooker, S.
\newblock The state of sparsity in deep neural networks.
\newblock \emph{arXiv preprint arXiv:1902.09574}, 2019.

\bibitem[Gray et~al.(2019)Gray, Crowley, and Storkey]{gray2019separable}
Gray, G., Crowley, E.~J., and Storkey, A.
\newblock Separable layers enable structured efficient linear substitutions.
\newblock \emph{arXiv preprint arXiv:1906.00859}, 2019.

\bibitem[Han et~al.(2015)Han, Pool, Tran, and Dally]{han2015learning}
Han, S., Pool, J., Tran, J., and Dally, W.~J.
\newblock Learning both weights and connections for efficient neural network.
\newblock In \emph{Advances in Neural Information Processing Systems}, 2015.

\bibitem[Han et~al.(2016)Han, Mao, and Dally]{han2016deep}
Han, S., Mao, H., and Dally, W.~J.
\newblock Deep compression: Compressing deep neural networks with pruning,
  trained quantization and {H}uffman coding.
\newblock In \emph{International Conference on Learning Representations}, 2016.

\bibitem[Hanson \& Pratt(1989)Hanson and Pratt]{hanson1989comparing}
Hanson, S.~J. and Pratt, L.~Y.
\newblock Comparing biases for minimal network construction with
  back-propagation.
\newblock In \emph{Advances in Neural Information Processing Systems}, 1989.

\bibitem[He et~al.(2016)He, Zhang, Ren, and Sun]{he2016deep}
He, K., Zhang, X., Ren, S., and Sun, J.
\newblock Deep residual learning for image recognition.
\newblock In \emph{Proceedings of the IEEE Conference on Computer Vision and
  Pattern Recognition}, 2016.

\bibitem[He et~al.(2018{\natexlab{a}})He, Girshick, and
  Doll{\'a}r]{he2018rethinking}
He, K., Girshick, R., and Doll{\'a}r, P.
\newblock Rethinking {I}mage{N}et pre-training.
\newblock \emph{arXiv preprint arXiv:1811.08883}, 2018{\natexlab{a}}.

\bibitem[He et~al.(2017)He, Zhang, and Sun]{he2017channel}
He, Y., Zhang, X., and Sun, J.
\newblock Channel pruning for accelerating very deep neural networks.
\newblock In \emph{International Conference on Computer Vision}, 2017.

\bibitem[He et~al.(2018{\natexlab{b}})He, Lin, Liu, Wang, Li, and
  Han]{he2018amc}
He, Y., Lin, J., Liu, Z., Wang, H., Li, L.-J., and Han, S.
\newblock {AMC}: Auto{ML} for model compression and acceleration on mobile
  devices.
\newblock In \emph{European Conference on Computer Vision}, 2018{\natexlab{b}}.

\bibitem[Hoffer et~al.(2017)Hoffer, Hubara, and Soudry]{hoffer2017train}
Hoffer, E., Hubara, I., and Soudry, D.
\newblock Train longer, generalize better: closing the generalization gap in
  large batch training of neural networks.
\newblock In \emph{Advances in Neural Information Processing Systems}, 2017.

\bibitem[Huang et~al.(2016)Huang, Sun, Liu, Sedra, and
  Weinberger]{huang2016deep}
Huang, G., Sun, Y., Liu, Z., Sedra, D., and Weinberger, K.~Q.
\newblock Deep networks with stochastic depth.
\newblock In \emph{European Conference on Computer Vision}, 2016.

\bibitem[Huang et~al.(2017)Huang, Liu, van~der Maaten, and
  Weinberger]{huang2017densely}
Huang, G., Liu, Z., van~der Maaten, L., and Weinberger, K.~Q.
\newblock Densely connected convolutional networks.
\newblock In \emph{Proceedings of the IEEE Conference on Computer Vision and
  Pattern Recognition}, 2017.

\bibitem[Huang et~al.(2018)Huang, Liu, van~der Maaten, and
  Weinberger]{huang2018condensenet}
Huang, G., Liu, S., van~der Maaten, L., and Weinberger, K.~Q.
\newblock Condense{N}et: An efficient densenet using learned group
  convolutions.
\newblock In \emph{Proceedings of the IEEE Conference on Computer Vision and
  Pattern Recognition}, 2018.

\bibitem[Ioannou et~al.(2017)Ioannou, Robertson, Cipolla, and
  Criminisi]{ioannou2017deep}
Ioannou, Y., Robertson, D., Cipolla, R., and Criminisi, A.
\newblock Deep roots: Improving {CNN} efficiency with hierarchical filter
  groups.
\newblock In \emph{Proceedings of the IEEE Conference on Computer Vision and
  Pattern Recognition}, 2017.

\bibitem[Ioffe \& Szegedy(2015)Ioffe and Szegedy]{ioffe2015batch}
Ioffe, S. and Szegedy, C.
\newblock Batch normalization: Accelerating deep network training by reducing
  internal covariate shift.
\newblock In \emph{International Conference on Machine Learning}, 2015.

\bibitem[Jastrz{\k{e}}bski et~al.(2018)Jastrz{\k{e}}bski, Arpit, Ballas, Verma,
  Che, and Bengio]{jastrzkebski2018residual}
Jastrz{\k{e}}bski, S., Arpit, D., Ballas, N., Verma, V., Che, T., and Bengio,
  Y.
\newblock Residual connections encourage iterative inference.
\newblock In \emph{International Conference on Learning Representations}, 2018.

\bibitem[Jastrz{\k{e}}bski et~al.(2019)Jastrz{\k{e}}bski, Kenton, Ballas,
  Fischer, Bengio, and Storkey]{jastrzkebski2019relation}
Jastrz{\k{e}}bski, S., Kenton, Z., Ballas, N., Fischer, A., Bengio, Y., and
  Storkey, A.
\newblock On the relation between the sharpest directions of {DNN} loss and the
  {SGD} step length.
\newblock In \emph{International Conference on Learning Representations}, 2019.

\bibitem[Jouppi et~al.(2017)Jouppi, Young, Patil, Patterson, Agrawal, Bajwa,
  Bates, Bhatia, Boden, Borchers, et~al.]{jouppi2017datacenter}
Jouppi, N.~P., Young, C., Patil, N., Patterson, D., Agrawal, G., Bajwa, R.,
  Bates, S., Bhatia, S., Boden, N., Borchers, A., et~al.
\newblock In-datacenter performance analysis of a tensor processing unit.
\newblock In \emph{International Symposium on Computer Architecture}, 2017.

\bibitem[Krizhevsky(2009)]{krizhevsky2009learning}
Krizhevsky, A.
\newblock Learning multiple layers of features from tiny images.
\newblock Master's thesis, University of Toronto, 2009.

\bibitem[Krizhevsky et~al.(2012)Krizhevsky, Sutskever, and
  Hinton]{krizhevsky2012imagenet}
Krizhevsky, A., Sutskever, I., and Hinton, G.
\newblock Image{N}et classification with deep convolutional neural networks.
\newblock In \emph{Advances in Neural Information Processing Systems}, 2012.

\bibitem[LeCun et~al.(1989)LeCun, Denker, and Solla]{lecun1989optimal}
LeCun, Y., Denker, J.~S., and Solla, S.~A.
\newblock Optimal brain damage.
\newblock In \emph{Advances in Neural Information Processing Systems}, 1989.

\bibitem[Lee et~al.(2019)Lee, Ajanthan, and Torr]{lee2019snip}
Lee, N., Ajanthan, T., and Torr, P. H.~S.
\newblock {SNIP}: Single-shot network pruning based on connection sensitivity.
\newblock In \emph{International Conference on Learning Representations}, 2019.

\bibitem[Li et~al.(2016)Li, Kadav, Durdanovic, Samet, and Graf]{li2016pruning}
Li, H., Kadav, A., Durdanovic, I., Samet, H., and Graf, H.~P.
\newblock Pruning filters for efficient convnets.
\newblock In \emph{International Conference on Learning Representations}, 2016.

\bibitem[Liu et~al.(2019{\natexlab{a}})Liu, Simonyan, and Yang]{liu2019darts}
Liu, H., Simonyan, K., and Yang, Y.
\newblock {DARTS}: Differentiable architecture search.
\newblock In \emph{International Conference on Learning Representations},
  2019{\natexlab{a}}.

\bibitem[Liu et~al.(2017)Liu, Li, Shen, Huang, Yan, and Zhang]{liu2017learning}
Liu, Z., Li, J., Shen, Z., Huang, G., Yan, S., and Zhang, C.
\newblock Learning efficient convolutional networks through network slimming.
\newblock In \emph{International Conference on Computer Vision}, 2017.

\bibitem[Liu et~al.(2019{\natexlab{b}})Liu, Sun, Zhou, Huang, and
  Darrell]{liu2019rethinking}
Liu, Z., Sun, M., Zhou, T., Huang, G., and Darrell, T.
\newblock Rethinking the value of network pruning.
\newblock In \emph{International Conference on Learning Representations},
  2019{\natexlab{b}}.

\bibitem[Louizos et~al.(2017)Louizos, Ullrich, and
  Welling]{louizos2017bayesian}
Louizos, C., Ullrich, K., and Welling, M.
\newblock Bayesian compression for deep learning.
\newblock In \emph{Advances in Neural Information Processing Systems}, 2017.

\bibitem[Molchanov et~al.(2017{\natexlab{a}})Molchanov, Ashukha, and
  Vetrov]{molchanov2017variational}
Molchanov, D., Ashukha, A., and Vetrov, D.
\newblock Variational dropout sparsifies deep neural networks.
\newblock In \emph{International Conference on Machine Learning},
  2017{\natexlab{a}}.

\bibitem[Molchanov et~al.(2017{\natexlab{b}})Molchanov, Tyree, Karras, Aila,
  and Kautz]{molchanov2017pruning}
Molchanov, P., Tyree, S., Karras, T., Aila, T., and Kautz, J.
\newblock Pruning convolutional neural networks for resource efficient
  inference.
\newblock In \emph{International Conference on Learning Representations},
  2017{\natexlab{b}}.

\bibitem[Parashar et~al.(2017)Parashar, Rhu, Mukkara, Puglielli, Venkatesan,
  Khailany, Emer, Keckler, and Dally]{parashar2017scnn}
Parashar, A., Rhu, M., Mukkara, A., Puglielli, A., Venkatesan, R., Khailany,
  B., Emer, J., Keckler, S.~W., and Dally, W.~J.
\newblock {SCNN}: An accelerator for compressed-sparse convolutional neural
  networks.
\newblock In \emph{International Symposium on Computer Architecture}, 2017.

\bibitem[Russakovsky et~al.(2015)Russakovsky, Deng, Su, Krause, Satheesh, Ma,
  Huang, Karpathy, Khosla, Bernstein, Berg, and
  Fei-Fei]{russakovsky2015imagenet}
Russakovsky, O., Deng, J., Su, H., Krause, J., Satheesh, S., Ma, S., Huang, Z.,
  Karpathy, A., Khosla, A., Bernstein, M., Berg, A.~C., and Fei-Fei, L.
\newblock Image{N}et large scale visual recognition challenge.
\newblock \emph{Int. Journal of Computer Vision (IJCV)}, 115\penalty0
  (3):\penalty0 211--252, 2015.
\newblock \doi{10.1007/s11263-015-0816-y}.

\bibitem[Scardapane et~al.(2017)Scardapane, Comminiello, Hussain, and
  Uncini]{scardapane2017group}
Scardapane, S., Comminiello, D., Hussain, A., and Uncini, A.
\newblock Group sparse regularization for deep neural networks.
\newblock \emph{Neurocomputing}, 241:\penalty0 81--89, 2017.

\bibitem[Simonyan \& Zisserman(2015)Simonyan and Zisserman]{simonyan2015deep}
Simonyan, K. and Zisserman, A.
\newblock Very deep convolutional networks for large-scale image recognition.
\newblock In \emph{International Conference on Learning Representations}, 2015.

\bibitem[Springenberg et~al.(2015)Springenberg, Dosovitskiy, Brox, and
  Riedmiller]{springenberg2015striving}
Springenberg, J.~T., Dosovitskiy, A., Brox, T., and Riedmiller, M.
\newblock Striving for simplicity: The all convolutional net.
\newblock In \emph{ICLR Workshop Track}, 2015.

\bibitem[Sze et~al.(2017)Sze, Chen, Yang, and Emer]{sze2017efficient}
Sze, V., Chen, Y.-H., Yang, T.-J., and Emer, J.~S.
\newblock Efficient processing of deep neural networks: A tutorial and survey.
\newblock \emph{Proceedings of the IEEE}, 105\penalty0 (12):\penalty0
  2295--2329, 2017.

\bibitem[Theis et~al.(2018)Theis, Korshunova, Tejani, and
  Husz{\'a}r]{theis2018faster}
Theis, L., Korshunova, I., Tejani, A., and Husz{\'a}r, F.
\newblock Faster gaze prediction with dense networks and {F}isher pruning.
\newblock \emph{arXiv preprint arXiv:1801.05787}, 2018.

\bibitem[Turner et~al.(2018)Turner, Cano, Radu, Crowley, O'Boyle, and
  Storkey]{turner2018characterising}
Turner, J., Cano, J., Radu, V., Crowley, E.~J., O'Boyle, M., and Storkey, A.
\newblock Characterising across stack optimisations for deep convolutional
  neural networks.
\newblock In \emph{IEEE International Symposium on Workload Characterization},
  2018.

\bibitem[Veit et~al.(2016)Veit, Wilber, and Belongie]{veit2016residual}
Veit, A., Wilber, M.~J., and Belongie, S.
\newblock Residual networks behave like ensembles of relatively shallow
  networks.
\newblock In \emph{Advances in Neural Information Processing Systems}, 2016.

\bibitem[Yang et~al.(2018)Yang, Howard, Chen, Zhang, Go, Sze, and
  Adam]{yang2018netadapt}
Yang, T.-J., Howard, A., Chen, B., Zhang, X., Go, A., Sze, V., and Adam, H.
\newblock Net{A}dapt: Platform-aware neural network adaptation for mobile
  applications.
\newblock In \emph{European Conference on Computer Vision}, 2018.

\bibitem[Ye et~al.(2018)Ye, Lu, Lin, and Wang]{ye2018rethinking}
Ye, J., Lu, X., Lin, Z., and Wang, J.~Z.
\newblock Rethinking the smaller-norm-less-informative assumption in channel
  pruning of convolution layers.
\newblock In \emph{International Conference on Learning Representations}, 2018.

\bibitem[Yu \& Koltun(2016)Yu and Koltun]{dilated2016}
Yu, F. and Koltun, V.
\newblock Multi-scale context aggregation by dilated convolutions.
\newblock In \emph{International Conference on Learning Representations}, 2016.

\bibitem[Zagoruyko \& Komodakis(2016)Zagoruyko and
  Komodakis]{zagoruyko2016wide}
Zagoruyko, S. and Komodakis, N.
\newblock Wide residual networks.
\newblock In \emph{British Machine Vision Conference}, 2016.

\bibitem[Zoph et~al.(2018)Zoph, Vasudevan, Shlens, and Le]{zoph2018learning}
Zoph, B., Vasudevan, V., Shlens, J., and Le, Q.~V.
\newblock Learning transferable architectures for scalable image recognition.
\newblock In \emph{Proceedings of the IEEE Conference on Computer Vision and
  Pattern Recognition}, 2018.

\end{thebibliography}
\bibliographystyle{icml2019}
\end{document}